\documentclass[10pt,twocolumn,letterpaper]{article}

\usepackage[T1]{fontenc}
\PassOptionsToPackage{table}{xcolor}
\usepackage[pagenumbers]{wacv} 

\definecolor{wacvblue}{rgb}{0.21,0.49,0.74}
\usepackage[pagebackref,breaklinks,colorlinks,allcolors=wacvblue]{hyperref}

\def\wacvPaperID{690} 
\def\confName{WACV}
\def\confYear{2027}

\title{CLEAR: Cooperative Latent Flow for Visual Counterfactual Explanations with Accurate Region Preservation}

\author{\begin{tabular}{@{}c@{\quad}c@{\quad}c@{}}
Yan Zeng & Changlu Guo\thanks{Changlu and Oskar contributed equally to this work.} & Oskar Kristoffersen\footnotemark[1]\\
Anders Nymark Christensen & Morten Rieger Hannemose & Anders Bjorholm Dahl
\end{tabular}\\
Technical University of Denmark\\
{\tt\small \{yanze,chagu,ofhkr,anym,mohan,abda\}@dtu.dk}
}

\usepackage{multirow}
\usepackage{booktabs}
\usepackage{graphicx}
\usepackage{tabularx}
\usepackage{array}

\begin{document}
\maketitle

\begin{abstract}
Visual counterfactual explanations (VCEs) should alter a classifier's
prediction while remaining realistic and preserving decision-irrelevant
content. Image-space methods provide high-quality counterfactuals but are
computationally costly, whereas latent-space methods are more efficient but
often less faithful. We identify two sources responsible for this gap. First, latent
generation recovers a natural sample from a perturbed representation, whereas
classifier guidance pushes it toward a target class, so their update
directions may oppose. Second, VAE reconstruction can alter regions that
should remain unchanged.
We introduce CLEAR, a latent Flow Matching framework for VCEs. CLEAR uses
Flow Matching's explicit instantaneous velocity as a reference for the
generative direction to decompose classifier guidance into
\textit{cooperative} and \textit{conflicting} components. It preserves the
cooperative component and adaptively attenuates only the conflicting one
according to conflict strength. We further analyze conflict frequency and
strength throughout generation to reveal how classifier guidance interacts
with the generative process. CLEAR also dynamically identifies editable
regions, constrains latent modifications, and directly preserves factual
content outside them to reduce VAE-induced changes. Across five bidirectional tasks on two datasets, CLEAR retains latent-space
efficiency while achieving counterfactual quality comparable to or better
than strong image-space baselines, with a $2.5\times$ speedup over the
strongest baseline.
Project page: \url{https://yan-zen9.github.io/CLEAR/}
\end{abstract}    
\section{Introduction}
\label{sec:intro}

Visual counterfactual explanations (VCEs) address a question: \emph{what meaningful change to an input image would alter a classifier's
prediction?}~\cite{wachter2017counterfactual,goyal2019counterfactual}
Unlike attribution methods that highlight potentially important regions,
VCEs show the visual changes needed to produce an alternative
prediction~\cite{
goyal2019counterfactual,
simonyan2013deep,
fong2017interpretable,
selvaraju2017gradcam}.
A reliable counterfactual should change the classifier prediction
while remaining realistic and preserving content unrelated to the target
decision~\cite{
jeanneret2023adversarial,
melistas2024benchmarking,
wang2020scout}. Diffusion models have improved VCE quality by keeping counterfactuals close
to the natural image distribution~\cite{
augustin2022diffusion,
jeanneret2022diffusion,
jeanneret2023adversarial,
augustin2024digin}.
However, image-space methods repeatedly perform generation and classifier
guidance in high-dimensional pixel space, leading to substantial
computational and memory costs~\cite{weng2024fast,guo2026maskdime}.
Latent generation provides an efficient alternative by operating on compact
representations~\cite{
luu2025visual,
farid2023latent,
lang2021explaining,
cohen2021gifsplanation}, but often at the cost of counterfactual fidelity.
The challenge is to retain latent-space VCE efficiency while approaching
the fidelity of image-space methods.

\begin{figure}[t]
\centering
\includegraphics[width=\columnwidth]{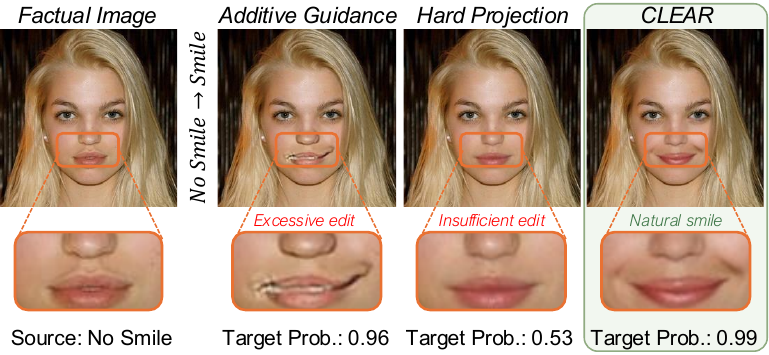}
\caption{
\textbf{Motivation for Cooperative Flow Guidance.}
Additive guidance may cause excessive editing, while hard projection can lead to insufficient editing. CLEAR adaptively regulates conflicting guidance to produce a natural and effective counterfactual.
}
\label{fig:diff}
\end{figure}

One key challenge is aligning classifier guidance with the generative
process. Latent generation seeks to recover a natural sample from a perturbed
representation, whereas classifier guidance pushes the sample toward a target
class. During counterfactual generation, these objectives can produce
opposing update directions. Importantly, such conflict does not mean that the
entire classifier guidance is harmful: a single signal can contain both
\textit{cooperative} and \textit{conflicting} components. Treating the whole
guidance uniformly may either retain harmful conflict or discard useful
information. CLEAR uses the explicit instantaneous velocity provided by Flow
Matching~\cite{lipman2022flow} as a reference for the current generative
direction. Unlike the sampler-dependent updates in
DDPMs~\cite{ho2020denoising}, this explicit direction allows CLEAR to
decompose the components of classifier guidance with respect to the current generative flow.
CLEAR preserves the cooperative component and adaptively attenuates only the
conflicting component according to conflict strength. As shown in
Fig.~\ref{fig:diff}, this avoids both excessive editing from fully retained
conflicting guidance, and insufficient modification from removing it entirely.
We further analyze directional conflicts throughout generation by examining
their frequency, strength, and effect on the final counterfactuals, showing
how classifier guidance interacts with the generative process.

A second challenge comes from VAE reconstruction, which can modify regions
that should remain unchanged~\cite{rombach2022high,luu2025visual}. CLEAR
therefore identifies editable regions dynamically and maps them to latent
resolution to constrain classifier-guided modifications throughout
generation. Outside this support, the latent trajectory follows the original flow, and the
final content outside the editable support is directly preserved in image
space, preventing VAE reconstruction and latent editing leakage from
affecting pixels outside the estimated support. Our main contributions are:
\begin{itemize}
\item We propose Cooperative Flow Guidance, which uses the explicit
velocity field of Flow Matching for component-wise, conflict-aware
regulation of classifier guidance during counterfactual generation.

\item We systematically analyze directional conflicts during counterfactual generation to clarify the interaction between classifier guidance and the generative process.

\item We introduce Adaptive Support Preservation (ASP), which dynamically localizes editable regions and preserves decision-irrelevant factual content across latent and image spaces.

\item On two CelebA-HQ attributes and three ImageNet class pairs, CLEAR improves FID and sFID over the evaluated latent-space baselines; the CelebA-HQ Smile benchmark demonstrates a favorable quality--efficiency trade-off against image-space methods.

\end{itemize}
\section{Related Work}
\label{sec:related}

\textbf{Visual counterfactual explanations and image-space generation.}
Early visual counterfactual methods explored region-based, generative,
and semantic formulations for identifying classifier-relevant
changes~\cite{goyal2019counterfactual,wang2020scout,
lang2021explaining,kim2023grounding}.
More recent work has further investigated fine-grained localization and
classifier-faithful counterfactual explanations~\cite{
zhang2025finegrained,chowdhury2025mirror}.
Diffusion models have become a strong foundation for generating
high-quality visual counterfactual explanations~\cite{
augustin2022diffusion,jeanneret2022diffusion,
jeanneret2023adversarial,augustin2024digin}.
DiME~\cite{jeanneret2022diffusion} applies classifier guidance throughout
image-space diffusion sampling, but requires costly iterative generation
and gradient computation.
FastDiME~\cite{weng2024fast} reduces this computational burden through
more efficient counterfactual sampling, while
MaskDiME~\cite{guo2026maskdime} further introduces adaptive masking to
dynamically focus generation on decision-relevant regions, improving both
efficiency and spatial localization.
Despite these advances, generation and classifier guidance are still
performed primarily in high-dimensional image space, motivating more
efficient latent-space alternatives.

\textbf{Latent visual counterfactual explanations.}
Earlier approaches explored counterfactual editing through compact
generative representations and latent traversals~\cite{
cohen2021gifsplanation,lang2021explaining}.
LDCE~\cite{farid2023latent} and ECED~\cite{luu2025visual}
are the closest latent VCE methods.
LDCE improves efficiency through latent diffusion and introduces consensus
guidance, which filters classifier gradients according to their agreement
with the diffusion model's implicit classifier.
This recognizes that classifier guidance may not always agree with the
semantic signal provided by the generator, but it compares two
discriminative signals and performs patch-level filtering rather than
explicitly relating classifier guidance to the current generative
direction or separating its useful and conflicting components.
ECED~\cite{luu2025visual} considers spatial localization and
background preservation by restricting edits through visual explanations.
To reduce VAE-induced reconstruction changes, it optimizes
the background latent representation and fine-tunes the VAE decoder for
each input, introducing test-time optimization that partially
offsets the efficiency advantage of latent generation.
Thus, LDCE primarily addresses classifier-gradient reliability, whereas
ECED improves spatial and reconstruction fidelity; neither jointly
resolves component-wise guidance conflict with the current generative
direction and efficient preservation of decision-irrelevant content
within a unified framework.

\textbf{Classifier guidance and flow matching.}
Classifier guidance steers generative models using discriminative
signals~\cite{dhariwal2021diffusion}, while subsequent works extend
guidance to external objectives~\cite{
bansal2023universal,go2023towards,wallace2023end}
or adaptively control its behavior during generation~\cite{
kwon2025tcfg,gao2026c2fg,wang2026cfgctrl}.
However, these adaptive schemes mainly target general image generation
and classifier-free guidance~\cite{ho2022classifierfree}, rather than
the component-wise interaction between external classifier guidance
and the current generative direction in latent VCEs.
Flow Matching~\cite{lipman2022flow} explicitly parameterizes an
instantaneous velocity field, providing a natural reference for this
directional relationship, and recent work has studied guidance in
flow-based generation~\cite{feng2025guidance}.
LeapFactual~\cite{cao2025leapfactual} further applies conditional flow
matching to reliable model-agnostic counterfactual generation.
In contrast, CLEAR uses the instantaneous latent flow direction to
decompose external classifier guidance and selectively modulate only
its generation-conflicting component, while adaptively preserving
decision-irrelevant regions throughout counterfactual generation.

 \begin{figure*}[t]
    \centering
    \includegraphics[width=0.85\textwidth]{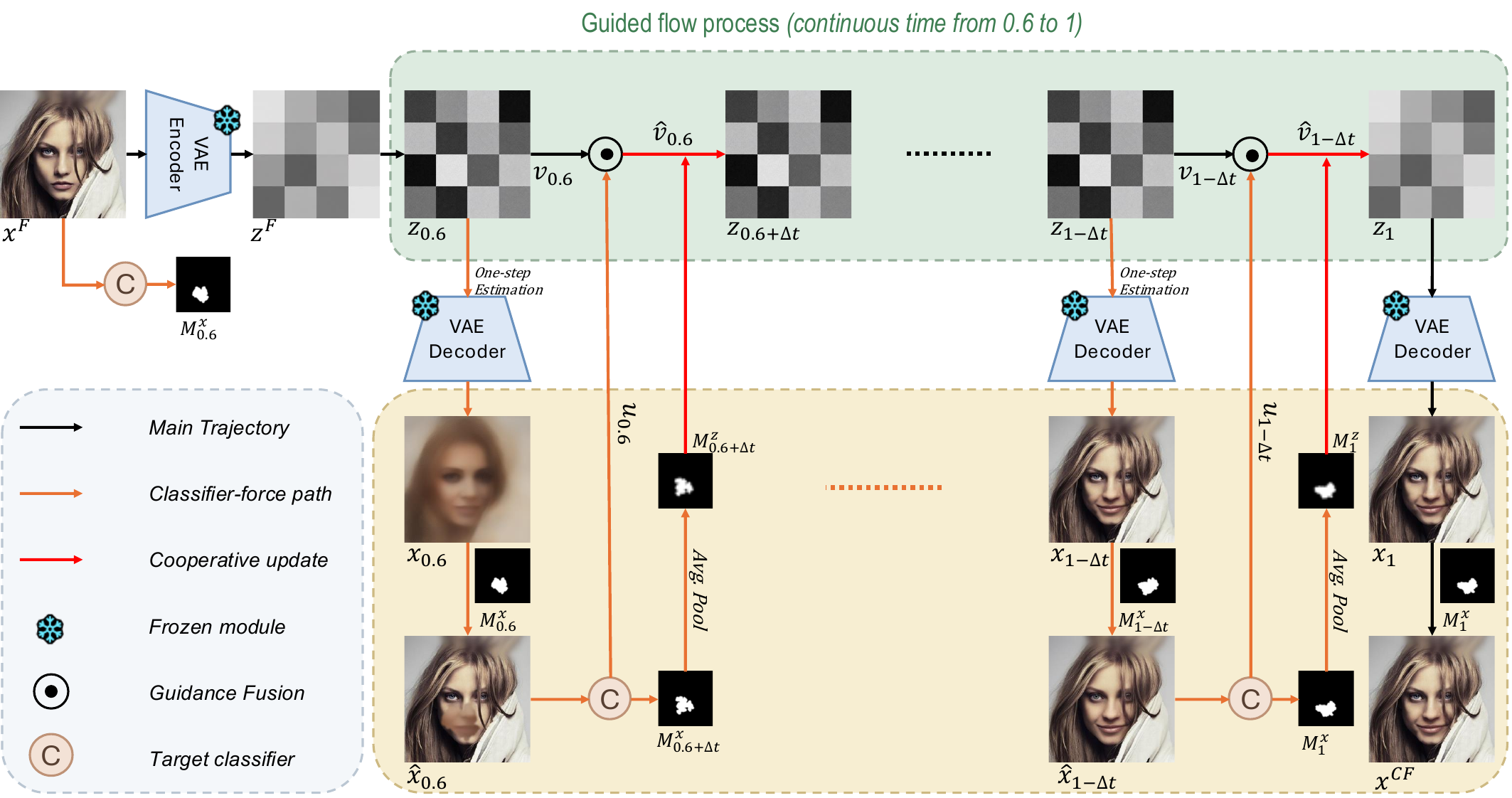}
    \caption{
    \textbf{Overview of CLEAR.}
    Starting from a partially noised factual latent, CLEAR evolves the
    counterfactual along a guided latent flow.
    At each timestep, the flow model provides the generative velocity
    and a one-step image estimate, from which the target classifier
    produces both the classifier force and the current decision-relevant
    support.
    The image-space support is mapped to the latent grid to constrain
    where classifier-driven modifications are introduced, while
    Cooperative Flow Guidance determines how the classifier force
    modifies the generative direction.
    Repeating these coupled updates yields a counterfactual that
    concentrates changes within the evolving editable support while
    preserving decision-irrelevant factual content.
    }
    \label{fig:arc}
\end{figure*}
\section{Method}
\label{sec:method}

\begin{figure*}[t]
    \centering
     \includegraphics[width=0.85\textwidth]{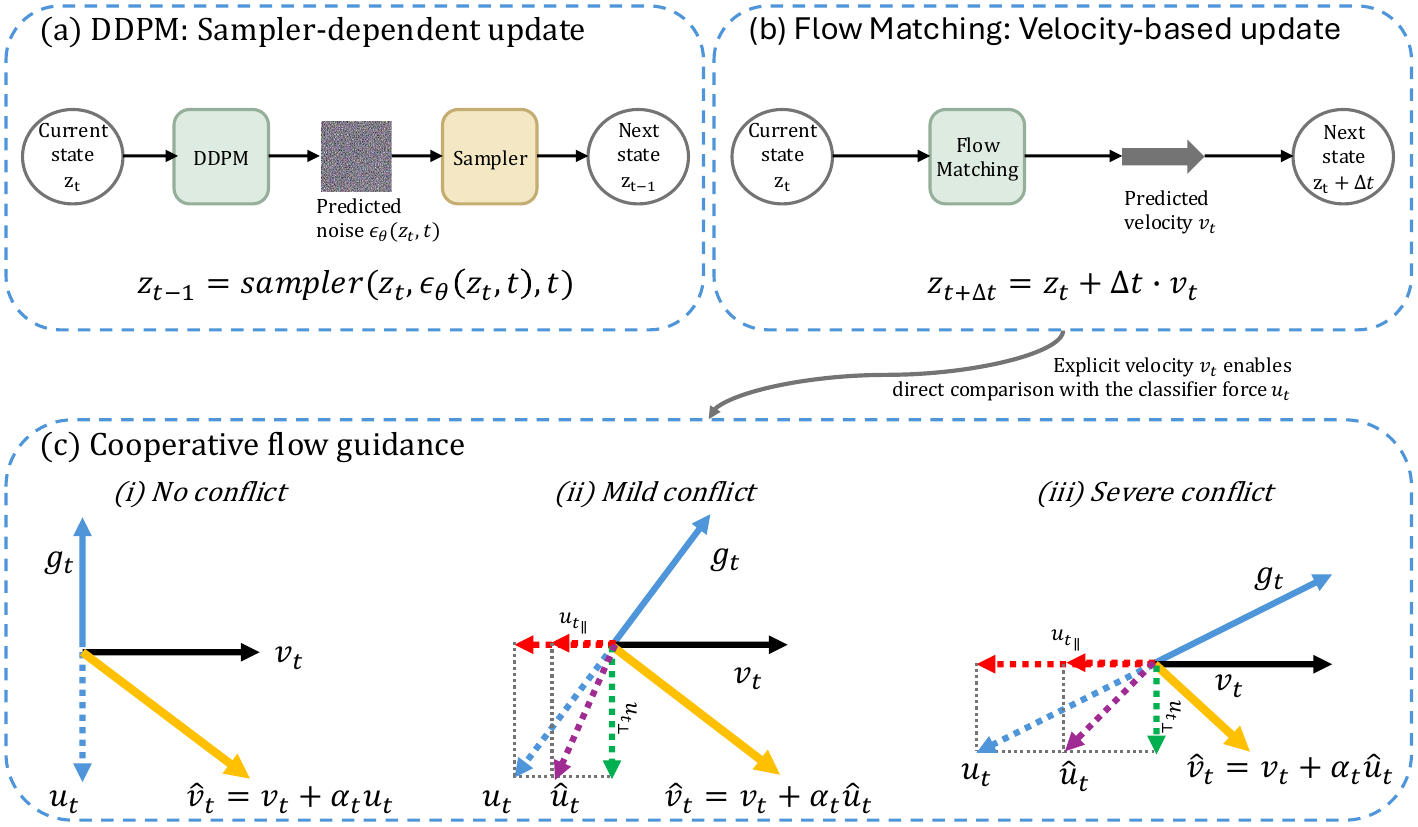}
    \caption{
    \textbf{From sampler-dependent denoising to cooperative flow guidance.}
    (a) In DDPMs, the predicted noise is converted into a state update through
    a chosen sampler, making the update direction sampler-dependent.
    (b) Flow Matching directly predicts the instantaneous velocity $v_t$,
    providing an explicit generative direction for comparison with the
    classifier force.
    (c) CLEAR reconciles the classifier force $u_t=-g_t$ according to its
    directional conflict with $v_t$. In the non-conflicting case (i.e. $\langle g_t,v_t\rangle \leq 0$), illustrated
    here by an orthogonal example, $u_t$ is retained. Under mild or severe
    conflict, CLEAR decomposes $u_t$ into parallel and orthogonal components
    and increasingly attenuates only the opposing component as the conflict
    strength grows, yielding the reconciled force $\hat{u}_t$ and guided
    velocity $\hat{v}_t=v_t+\alpha_t\hat{u}_t$.
    }
    \label{fig:coop}
\end{figure*}

\subsection{Latent Flow Setup and Classifier View}
\label{sec:latent_setup}

Figure~\ref{fig:arc} provides an overview of CLEAR, which combines
Cooperative Flow Guidance with Adaptive Support Preservation during
latent flow generation.

Let $E$ and $D$ denote the encoder and decoder of a frozen
VAE~\cite{rombach2022high}. Given a factual image $x^F$, we first obtain
its latent representation
\begin{equation}
z^F = E(x^F).
\end{equation}

Rather than generating the counterfactual from pure noise, CLEAR starts
from a partially noised version of the factual latent. Given Gaussian
noise $\epsilon\sim\mathcal{N}(0,I)$, the initial latent state is
\begin{equation}
z_{t_s}
=
t_s z^F + (1-t_s)\epsilon .
\label{eq:partial_noise}
\end{equation}
Under the Flow Matching~\cite{lipman2022flow} parameterization used in this work, $t=0$
corresponds to the noise endpoint and $t=1$ to the clean data endpoint.
Starting from an intermediate time $t_s$ retains factual semantic
structure while leaving sufficient freedom for counterfactual modification.

Starting from $z_{t_s}$, CLEAR evolves a single latent trajectory toward
$t=1$. At timestep $t$, the Flow Matching model predicts the
instantaneous velocity
\begin{equation}
v_t = v_\theta(z_t,t),
\label{eq:flow_velocity}
\end{equation}
which describes the current generative direction in latent space.
Under the linear Flow Matching parameterization, we obtain a one-step
estimate of the clean sample by extrapolating to the data endpoint and
decoding it through the frozen VAE:
\begin{equation}
x_t
=
D\!\left(
z_t + (1-t)v_t
\right).
\label{eq:clean_estimate}
\end{equation}
The resulting $x_t$ is an image-space estimate associated with the
current latent state and is used to obtain classifier feedback during
generation.

Let $f$ denote the frozen target classifier. To focus its feedback on
the current editable region, we construct the classifier view
\begin{equation}
\hat{x}_t
=
M_t^x \odot x_t
+
(1-M_t^x)\odot x^F,
\label{eq:classifier_view}
\end{equation}
where $M_t^x$ denotes the current editable support in image space.
Its adaptive construction is described in
Sec.~\ref{sec:adaptive_support}.
The classifier view $\hat{x}_t$ provides the common image-space interface
for the two core components of CLEAR: Cooperative Flow Guidance derives
the classifier force acting on the latent trajectory, while Adaptive
Support Preservation determines where counterfactual modification is
permitted.

\subsection{Cooperative Flow Guidance}
\label{sec:cooperative_guidance}

CLEAR uses the instantaneous Flow Matching velocity $v_t$ as a reference
to reconcile classifier guidance~\cite{dhariwal2021diffusion} with the current generative direction.
Using the classifier view $\hat{x}_t$, we define
\begin{equation}
\begin{aligned}
\mathcal{L}_{\mathrm{cls}}
&=
\operatorname{CE}\!\left(f(\hat{x}_t),y^{CF}\right),\\
g_t
&=
\nabla_{z_t}\mathcal{L}_{\mathrm{cls}},\\
u_t
&=
-g_t.
\end{aligned}
\label{eq:classifier_force}
\end{equation}
where $u_t$ is the classifier force applied to the latent trajectory and $\operatorname{CE}$ is the cross entropy.
The classifier gradient is backpropagated through the frozen VAE decoder
to the current latent state $z_t$, allowing image-space classifier
feedback to guide the latent trajectory.

As illustrated in Fig.~\ref{fig:coop}, the sign of
$\langle u_t,v_t\rangle$ indicates whether the classifier force is
cooperative, orthogonal, or conflicting with the current flow.
Under conflict, CLEAR decomposes $u_t$ into components parallel and
orthogonal to $v_t$:
\begin{equation}
\begin{aligned}
u_{t\parallel}
&=
\frac{\langle u_t,v_t\rangle}
{\|v_t\|_2^2+\varepsilon}v_t,\\
u_{t\perp}
&=
u_t-u_{t\parallel} .
\end{aligned}
\label{eq:force_decomposition}
\end{equation}
where $\varepsilon=10^{-8}$ is a small constant for numerical stability.
Here, $u_{t\parallel}$ contains the opposing component under conflict,
whereas $u_{t\perp}$ does not directly counteract the generative flow.

CLEAR adapts the attenuation to the degree of directional conflict as
\begin{equation}
\beta_t
=
\operatorname{clip}
\left(
\frac{-\cos(u_t,v_t)}{\tau},
0,1
\right),
\label{eq:conflict_attenuation}
\end{equation}
where $\tau$ controls conflict sensitivity. Thus, $\beta_t=0$ for
non-conflicting directions, and increases with the degree of conflict.
The reconciled classifier force is
\begin{equation}
\hat{u}_t
=
u_{t\perp}+(1-\beta_t)u_{t\parallel} .
\label{eq:reconciled_force}
\end{equation}
Hence, CLEAR attenuates only the opposing parallel component while
preserving non-conflicting classifier information.

Finally, we prevent classifier guidance from overwhelming the
generative flow by bounding its magnitude:
\begin{equation}
\alpha_t =
\min\left(
s,\,
\frac{\|v_t\|_2}{\|\hat{u}_t\|_2+\varepsilon}
\right),
\qquad
\hat{v}_t = v_t + \alpha_t \hat{u}_t .
\label{eq:cooperative_velocity}
\end{equation}
Here, $s$ is the maximum guidance scale.

\subsection{Adaptive Support Preservation}
\label{sec:adaptive_support}

VAE reconstruction may introduce changes outside regions intended for
counterfactual editing. CLEAR therefore dynamically estimates an editable
support and uses it to spatially constrain generation.

Let $M_t^x\in\{0,1\}^{H\times W}$ denote the image-space editable
support. We compute the saliency map using Grad-CAM~\cite{selvaraju2017gradcam} with respect to the
target class $y^{CF}$. The initial support is obtained from the factual
image $x^F$, while subsequent supports are updated from the current
classifier view $\hat{x}_t$. For an Euler step of size $\Delta t$,
\begin{equation}
\begin{aligned}
S_t
&=
\operatorname{GradCAM}\!\left(f,\hat{x}_t,y^{CF}\right),\\
M_{t+\Delta t}^x
&=
\operatorname{Dilate}\!\left(
\operatorname{LCC}\!\left(
\operatorname{Top}_{\rho}(S_t)
\right)\right).
\end{aligned}
\label{eq:adaptive_support}
\end{equation}
Here, $\operatorname{Top}_{\rho}$ retains the top $\rho$ fraction of
pixels ranked by target-class Grad-CAM response,
$\operatorname{LCC}$ extracts the largest connected component, and
dilation provides spatial margin around the resulting support.
The initial support $M_{t_s}^x$ is constructed analogously using
target-class Grad-CAM on $x^F$. The new support replaces the previous
one, allowing it to adapt throughout generation.

Since the support represents spatial location rather than image content,
we map it to the latent spatial resolution by average pooling,
\begin{equation}
M_t^z=\mathcal{P}(M_t^x),
\label{eq:latent_support}
\end{equation}
where $\mathcal{P}$ denotes average pooling. This maps the binary
image-space support to a soft latent gate
$M_t^z\in[0,1]^{h\times w}$, providing smooth transitions near support
boundaries.

The resulting latent gate restricts classifier-driven editing as
\begin{equation}
\begin{aligned}
v_t^{\mathrm{ASP}}
&=
M_{t+\Delta t}^z\odot\hat{v}_t
+
(1-M_{t+\Delta t}^z)\odot v_t,\\
z_{t+\Delta t}
&=
z_t+\Delta t\,v_t^{\mathrm{ASP}}.
\end{aligned}
\label{eq:adaptive_latent_update}
\end{equation}
Thus, cooperative guidance acts inside the editable support, while the
remaining region follows the original Flow Matching dynamics.

Finally, we remove residual reconstruction changes outside the final
support by composing the output in image space:
\begin{equation}
x^{CF}
=
M_1^x\odot x_1
+
(1-M_1^x)\odot x^F,
\label{eq:final_output}
\end{equation}
where $x_1=D(z_1)$ is the decoded final latent. This combines adaptive
spatial editing with exact factual-pixel preservation outside the final
support.
\section{Experiments}
\label{sec:experiments}

\subsection{Experimental Setup}
\label{sec:experimental_setup}

\textbf{Datasets and implementation.} We follow the ACE protocol~\cite{jeanneret2023adversarial} for data
splits, preprocessing, and classifier settings. Both
CelebA-HQ~\cite{liu2015deep,karras2017progressive} and
ImageNet~\cite{deng2009imagenet} are evaluated at $256\times256$
resolution. On CelebA-HQ, we consider the bidirectional \emph{Smile}
and \emph{Age} tasks using a DenseNet-121~\cite{huang2017densely}
classifier. On ImageNet, we use a ResNet-50~\cite{he2016deep}
classifier and evaluate sorrel$\leftrightarrow$zebra,
Persian cat$\leftrightarrow$Egyptian cat, and
cougar$\leftrightarrow$cheetah. Both datasets use the frozen \texttt{stabilityai/sd-vae-ft-mse}
VAE~\cite{rombach2022high} with publicly released pretrained
Latent Flow Matching (LFM) checkpoints~\cite{dao2023flow}.
We use the unconditional DiT-L/2 checkpoint for CelebA-HQ and the
class-conditional DiT-B/2 checkpoint for ImageNet, which is used
unconditionally during counterfactual inference.
All pretrained models remain frozen, and CLEAR requires no additional
training or fine-tuning. For all tasks, CLEAR uses $t_s=0.6$, $N=20$ uniform Euler steps ($\Delta t=0.02$), and a maximum guidance scale $s=8$. Cooperative Flow Guidance uses $\tau=0.5$.
Adaptive Support Preservation retains the top $5\%$, $12\%$, and
$30\%$ salient pixels for \emph{Smile}, \emph{Age}, and ImageNet,
respectively, with a dilation radius of $4$ for all tasks. The
image-space support is mapped to the latent grid using average pooling.
Except for the support ratio, all CLEAR hyperparameters are shared
across tasks. All counterfactual inference and efficiency experiments
are performed on the same AMD Instinct MI250X 64\,GB GPU, and the
reported efficiency measures counterfactual inference only.

\textbf{Evaluation metrics.}
Following MaskDiME~\cite{guo2026maskdime}, we measure computational efficiency using wall-clock inference time and peak GPU memory. Following ACE~\cite{jeanneret2023adversarial}, counterfactual validity is evaluated using flip rate (FR) and COUT, while FID~\cite{heusel2017gans} and sFID measure~\cite{nash2021generating}
 distributional realism. On CelebA-HQ, FVA and face similarity (FS)~\cite{cao2018vggface2} assess identity preservation, whereas MNAC and CD quantify non-target attribute changes and deviations from their natural co-variation. On ImageNet, $S^3$~\cite{chen2021exploring} measures semantic similarity between factual and counterfactual images.

\begin{table*}[t]
\centering
\small
\setlength{\tabcolsep}{2.5pt}
\renewcommand{\arraystretch}{0.95}
\caption{
Quantitative comparison on \textbf{CelebA-HQ} across the
\emph{Smile} and \emph{Age} tasks. Methods marked with $^\dagger$
operate in latent space. Best and second-best results are highlighted
in \textbf{bold} and \underline{underline}, respectively.
}
\label{tab:celeba_hq}

\resizebox{\textwidth}{!}{%
\begin{tabular}{l|cccccccc|cccccccc}
\toprule
& \multicolumn{8}{c|}{\textbf{Smile}}
& \multicolumn{8}{c}{\textbf{Age}} \\
\cmidrule(lr){2-9}
\cmidrule(lr){10-17}

\textbf{Method}
& FID$\downarrow$
& sFID$\downarrow$
& FVA$\uparrow$
& FS$\uparrow$
& MNAC$\downarrow$
& CD$\downarrow$
& COUT$\uparrow$
& FR(\%)$\uparrow$
& FID$\downarrow$
& sFID$\downarrow$
& FVA$\uparrow$
& FS$\uparrow$
& MNAC$\downarrow$
& CD$\downarrow$
& COUT$\uparrow$
& FR(\%)$\uparrow$ \\
\midrule

DiVE~\cite{rodriguez2021beyond}
& 107.0 & -- & 35.7 & -- & 7.41 & -- & -- & --
& 107.5 & -- & 32.3 & -- & 6.76 & -- & -- & -- \\

STEEX~\cite{jacob2022steex}
& 21.9 & -- & 97.6 & -- & 5.27 & -- & -- & --
& 26.8 & -- & 96.0 & -- & 5.63 & -- & -- & -- \\

DiME~\cite{jeanneret2022diffusion}
& 18.1 & 27.7 & 96.7 & 0.67 & 2.63
& \underline{1.82} & 0.65 & 97.0
& 18.7 & 27.8 & 95.0 & 0.66 & 2.10
& 4.29 & 0.56 & 97.0 \\

ACE $\ell_1$~\cite{jeanneret2023adversarial}
& 3.21 & 20.2 & \textbf{100.0} & 0.89 & 1.56
& 2.61 & 0.55 & 95.0
& 5.31 & 21.7 & \underline{99.6} & 0.81 & 1.53
& 5.40 & 0.40 & 95.0 \\

ACE $\ell_2$~\cite{jeanneret2023adversarial}
& 6.93 & 22.0 & \textbf{100.0} & 0.84 & 1.87
& 2.21 & 0.60 & 95.0
& 16.4 & 28.2 & \underline{99.6} & 0.77 & 1.92
& \underline{4.21} & 0.53 & 95.0 \\

TiME~\cite{jeanneret2024text}
& 10.98 & 23.8 & 96.6 & 0.79 & 2.97
& 2.32 & 0.63 & 97.1
& 20.9 & 32.9 & 79.3 & 0.63 & 4.19
& 4.29 & 0.31 & 89.9 \\

RCSB~\cite{sobieski2025rethinking}
& 3.04 & 20.0 & \textbf{100.0}
& \textbf{0.94} & \textbf{1.22}
& 3.22 & \underline{0.83} & 98.9
& 4.92 & 27.3 & \textbf{100.0}
& \textbf{0.96} & 1.47
& 5.16 & \textbf{0.80} & 99.4 \\

MaskDiME~\cite{guo2026maskdime}
& \underline{2.51} & \underline{18.1} & \textbf{100.0}
& \textbf{0.94} & 1.41
& 2.67 & 0.69 & \underline{99.4}
& \underline{4.43} & \underline{19.4} & \textbf{100.0}
& 0.88 & 1.82
& 4.67 & 0.63 & \underline{99.6} \\

LDCE-txt$^\dagger$~\cite{farid2023latent}
& 13.6 & 25.8 & \underline{99.1} & 0.76 & 2.44
& \textbf{1.68} & 0.34 & --
& 14.2 & 25.6 & 98.0 & 0.73 & 2.12
& \textbf{4.02} & 0.33 & -- \\

ECED$^\dagger$~\cite{luu2025visual}
& 7.56 & 24.9 & \textbf{100.0}
& 0.90 & 1.32
& 5.08 & \textbf{0.84} & --
& 12.4 & 31.6 & \textbf{100.0}
& \textbf{0.96} & \textbf{0.84}
& 4.93 & \underline{0.74} & -- \\

\rowcolor{gray!10}
CLEAR$^\dagger$
& \textbf{2.18} & \textbf{17.4} & \textbf{100.0}
& \textbf{0.94} & \underline{1.30}
& 2.28 & 0.65 & \textbf{99.6}
& \textbf{4.11} & \textbf{19.0} & \textbf{100.0}
& \underline{0.91} & \underline{1.38}
& 4.94 & 0.67 & \textbf{100.0} \\

\bottomrule
\end{tabular}%
}
\end{table*}

\subsection{Computational Efficiency}
\label{sec:efficiency}

\begin{figure}[t]
    \centering
    \includegraphics[width=\columnwidth]{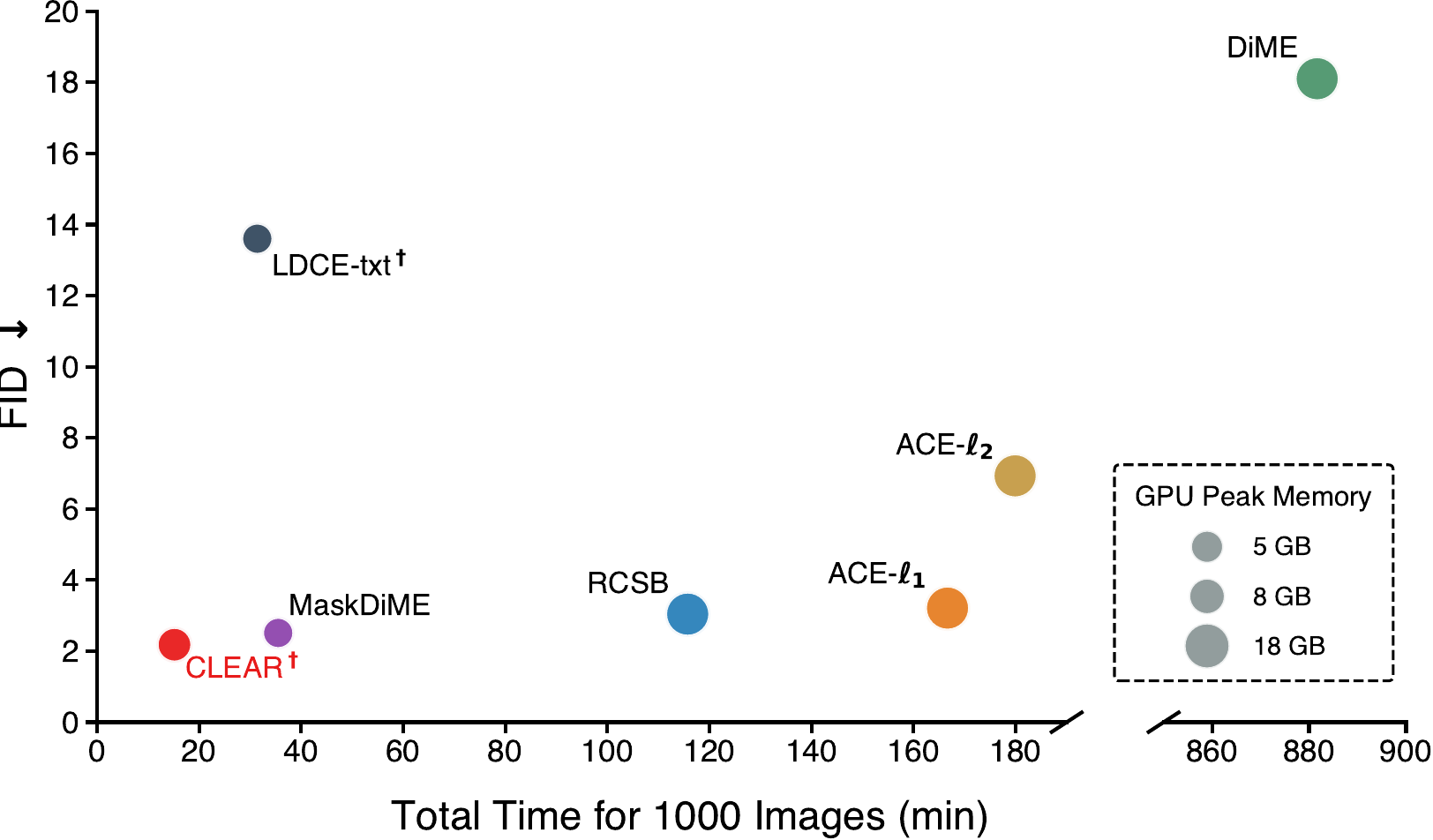}
    \caption{
    \textbf{Inference efficiency and counterfactual quality on
    CelebA-HQ \emph{Smile}.}
    All methods are evaluated on 1,000 images at $256\times256$ resolution
    with a batch size of 3. Runtime and peak GPU memory measure
    counterfactual inference only. $^\dagger$ denotes a latent-space method.
    }
    \label{fig:speed}
\end{figure}

Figure~\ref{fig:speed} compares the quality--efficiency trade-off of
CLEAR with representative image-space VCE methods and the
latent baseline LDCE-txt~\cite{farid2023latent}.
CLEAR achieves the lowest runtime and FID among the compared methods
while maintaining a low peak GPU memory footprint. Compared with the
image-space baselines, it provides $2.5\times$ and $58.1\times$
speedups over MaskDiME~\cite{guo2026maskdime} and
DiME~\cite{jeanneret2022diffusion}, respectively.
CLEAR is also substantially faster and achieves lower FID than
LDCE-txt, demonstrating a stronger quality--efficiency trade-off than
the existing latent baseline.

\begin{figure}[t]
    \centering
    \includegraphics[width=\columnwidth]{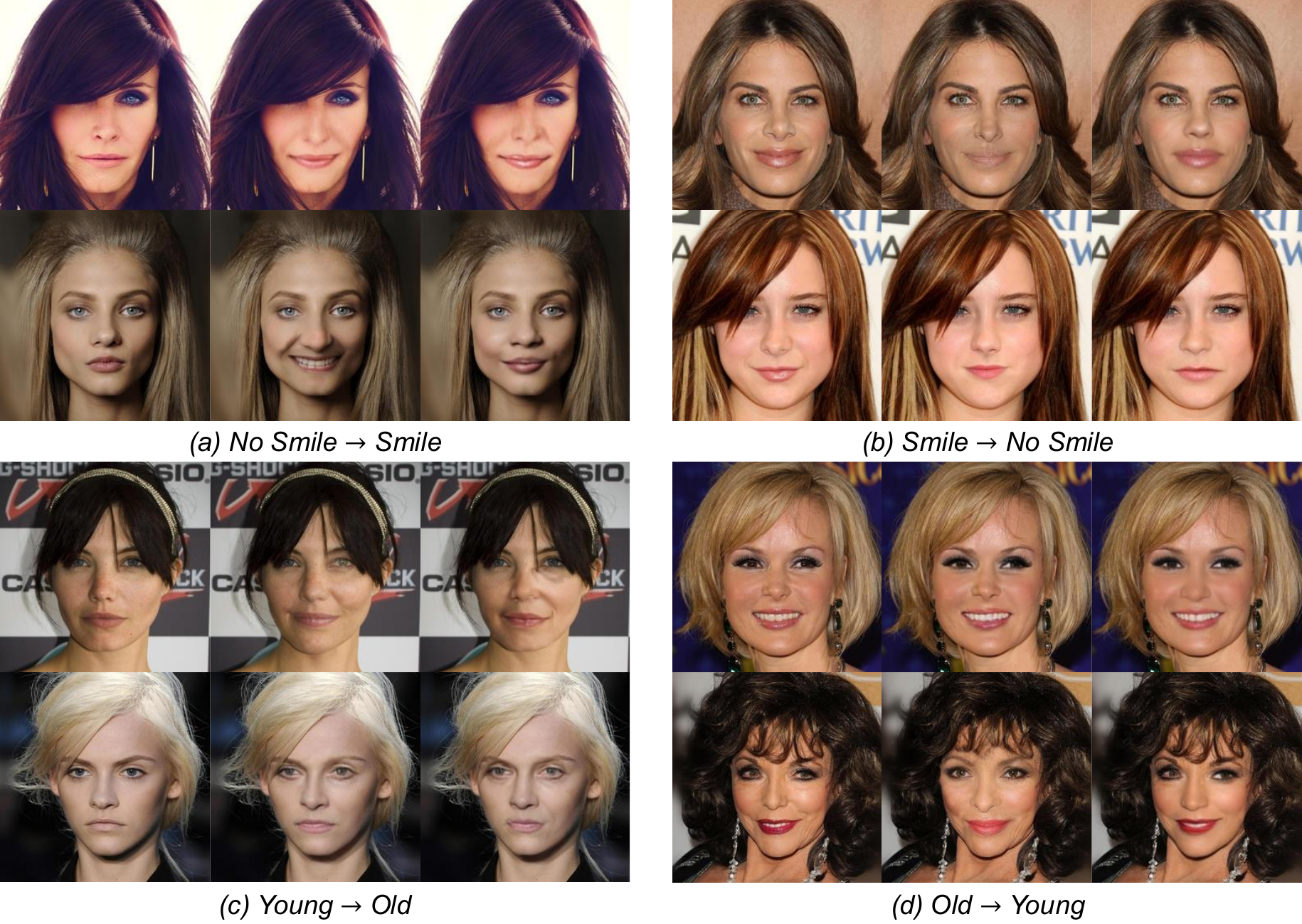}
    \caption{
    \textbf{Qualitative comparison on CelebA-HQ using DenseNet-121.}
    From left to right: the factual image and counterfactual images
    generated by MaskDiME and CLEAR.
    }
    \label{fig:comp_img}
\end{figure}

\subsection{Counterfactual Quality}
\label{sec:quality}

Having established the computational advantage of CLEAR, we next examine
whether this efficiency compromises counterfactual quality.

\textbf{CelebA-HQ.} As shown in Table~\ref{tab:celeba_hq}, CLEAR achieves the best FID and
sFID on both \emph{Smile} and \emph{Age}, while maintaining near-perfect
validity and strong identity preservation. Compared with existing latent VCE methods, CLEAR substantially improves realism and approaches or surpasses strong image-space baselines, narrowing the fidelity gap between latent- and image-space counterfactual generation while avoiding excessive non-target changes. The qualitative results in Fig.~\ref{fig:comp_img} further demonstrate localized edits with strong identity preservation.

\begin{table*}[t]
\centering
\small
\setlength{\tabcolsep}{2.4pt}
\renewcommand{\arraystretch}{0.95}
\caption{
Quantitative comparison on \textbf{ImageNet} across three semantic
class pairs. Methods marked with $^\dagger$ operate in latent space.
Best and second-best results are highlighted in \textbf{bold} and
\underline{underline}, respectively.
}
\label{tab:imagenetresults_wide}

\resizebox{\textwidth}{!}{%
\begin{tabular}{l|ccccc|ccccc|ccccc}
\toprule
\multirow{2}{*}{\textbf{Method}}
& \multicolumn{5}{c|}{\textbf{Sorrel -- Zebra}}
& \multicolumn{5}{c|}{\textbf{Persian Cat -- Egyptian Cat}}
& \multicolumn{5}{c}{\textbf{Cougar -- Cheetah}} \\
\cmidrule(lr){2-6}
\cmidrule(lr){7-11}
\cmidrule(lr){12-16}

& FID$\downarrow$
& sFID$\downarrow$
& $S^3\uparrow$
& COUT$\uparrow$
& FR(\%)$\uparrow$
& FID$\downarrow$
& sFID$\downarrow$
& $S^3\uparrow$
& COUT$\uparrow$
& FR(\%)$\uparrow$
& FID$\downarrow$
& sFID$\downarrow$
& $S^3\uparrow$
& COUT$\uparrow$
& FR(\%)$\uparrow$ \\
\midrule

DiME~\cite{jeanneret2022diffusion}
& 222.9 & 243.2 & 0.19 & $-0.31$ & 0.0
& 322.8 & 352.1 & 0.44 & $-0.05$ & 0.0
& 268.2 & 292.0 & 0.11 & $-0.16$ & 0.0 \\

ACE $\ell_1$~\cite{jeanneret2023adversarial}
& 84.5 & 122.7 & 0.92 & $-0.45$ & 47.0
& 93.6 & 156.7 & 0.85 & 0.25 & 85.0
& 70.2 & 100.5 & 0.91 & 0.02 & 77.0 \\

ACE $\ell_2$~\cite{jeanneret2023adversarial}
& 67.7 & 98.4 & 0.90 & $-0.25$ & 81.0
& 107.3 & 160.4 & 0.78 & 0.34 & 97.0
& 74.1 & 102.5 & 0.88 & 0.12 & 95.0 \\

FastDiME~\cite{weng2024fast}
& 96.5 & 103.5 & 0.22 & $-0.44$ & 14.0
& 193.6 & 207.1 & 0.10 & 0.01 & 20.0
& 133.0 & 141.1 & 0.12 & $-0.11$ & 18.0 \\

DVCE~\cite{augustin2022diffusion}
& 33.1 & 43.9 & 0.62 & $-0.21$ & 57.8
& 46.6 & 59.2 & 0.59 & 0.60 & 98.5
& 46.9 & 54.1 & 0.70 & 0.49 & 99.0 \\

RCSB~\cite{sobieski2025rethinking}
& \textbf{8.0} & \textbf{16.2} & 0.88 & 0.74 & 94.7
& \textbf{23.0} & \textbf{40.0} & 0.87 & \textbf{0.92}
& \textbf{100.0}
& \textbf{17.2} & \textbf{26.6} & 0.92 & \textbf{0.92}
& \textbf{100.0} \\

MaskDiME~\cite{guo2026maskdime}
& 32.5 & 39.7 & 0.94 & \textbf{0.85} & 99.7
& 36.2 & 50.1 & 0.87 & \underline{0.89} & \textbf{100.0}
& 37.4 & 46.3 & 0.93 & \textbf{0.92} & \textbf{100.0} \\

LDCE-cls$^\dagger$~\cite{farid2023latent}
& 84.2 & 107.2 & 0.78 & $-0.06$ & 88.0
& 102.7 & 140.7 & 0.63 & 0.52 & \underline{99.0}
& 71.0 & 91.8 & 0.62 & 0.51 & \textbf{100.0} \\

LDCE-txt$^\dagger$~\cite{farid2023latent}
& 82.4 & 107.2 & 0.71 & $-0.21$ & 81.0
& 121.7 & 162.4 & 0.61 & 0.56 & \underline{99.0}
& 91.2 & 117.0 & 0.59 & 0.34 & 98.0 \\

ECED-v$^\dagger$~\cite{luu2025visual}
& 33.0 & 45.2 & \underline{0.95} & \underline{0.77} & 98.0
& 27.4 & 53.5 & \textbf{0.98} & 0.85 & \textbf{100.0}
& 23.3 & 37.3 & \underline{0.95} & 0.83
& \underline{99.5} \\

ECED-r$^\dagger$~\cite{luu2025visual}
& 63.1 & 97.8 & 0.86 & 0.68 & \underline{98.5}
& 44.3 & 71.2 & 0.94 & 0.81 & \textbf{100.0}
& 47.0 & 76.6 & \textbf{0.96} & 0.82 & \textbf{100.0} \\

\rowcolor{gray!10}
CLEAR$^\dagger$
& \underline{24.3} & \underline{31.3} & \textbf{0.96}
& \underline{0.77} & \textbf{99.9}
& \underline{26.7} & \underline{42.1} & \textbf{0.98}
& 0.82 & \textbf{100.0}
& \underline{21.6} & \underline{35.5} & \textbf{0.96}
& \underline{0.84} & \textbf{100.0} \\

\bottomrule
\end{tabular}%
}
\end{table*}

\begin{figure}[t]
    \centering
    \includegraphics[width=\columnwidth]{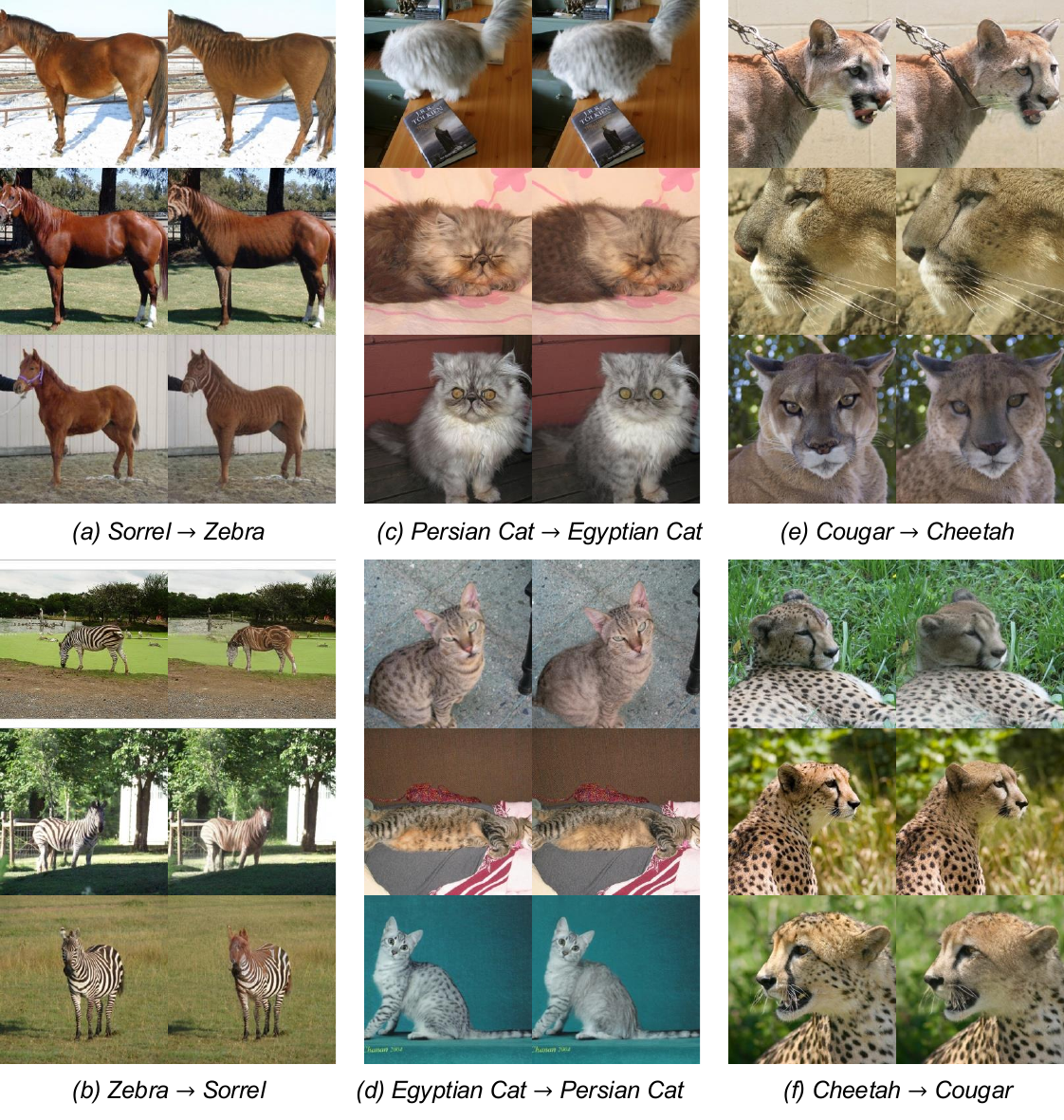}
    \caption{
    \textbf{Qualitative results on ImageNet using ResNet-50.}
    Each pair shows a factual image and the corresponding
    counterfactual generated by CLEAR for the paired target class.
    }
    \label{fig:vis_img}
\end{figure}

\textbf{ImageNet.} Table~\ref{tab:imagenetresults_wide} shows that CLEAR consistently
outperforms existing latent-space methods~\cite{farid2023latent,luu2025visual}
in FID and sFID across all three semantic class pairs, while remaining
competitive with the strongest image-space baselines. At the same time,
CLEAR achieves the best or tied-best $S^3$ with near-perfect flip rates.
These results show that CLEAR preserves its quality--efficiency advantage
beyond localized facial edits, extending to challenging global semantic
transformations. Figure~\ref{fig:vis_img} further shows that CLEAR handles challenging global semantic transformations.

\begin{figure*}[t]
\centering
\includegraphics[width=\textwidth]{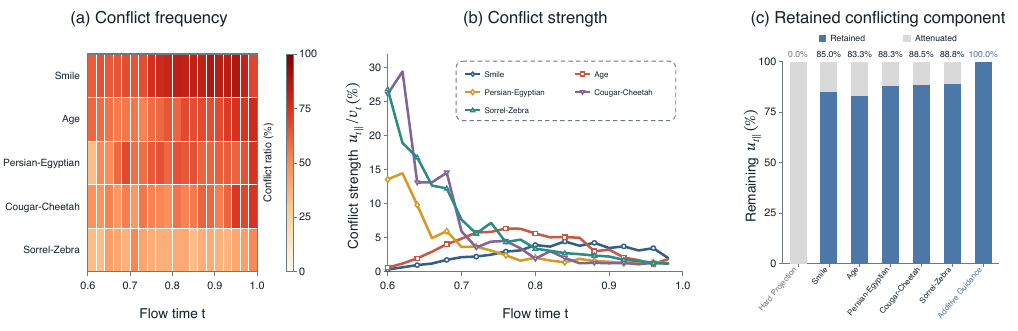}
\caption{
\textbf{Directional conflict analysis during generation.}
(a) Conflict frequency across Flow Matching timesteps.
(b) Conflict strength, measured by the relative magnitude
$\|u_{t\parallel}\|_2/\|v_t\|_2$.
(c) Percentage of the conflicting component retained after Cooperative Flow Guidance,
with Hard Projection and Additive Guidance corresponding to 0\% and 100\% retention, respectively.
}
\label{fig:conflict_heatmap}
\end{figure*}

\subsection{Directional Conflict Analysis}
\label{sec:conflict_analysis}

Figure~\ref{fig:conflict_heatmap} reveals the interaction between classifier
guidance and the current generative flow.
As shown in Fig.~\ref{fig:conflict_heatmap}(a), directional conflict occurs
frequently across all five bidirectional tasks and becomes more common toward
the end of the generation process.
This is counter-intuitive, since the classifier force is expected to
become more aligned with the generative direction as target-class semantics
emerge.
Instead, directional disagreement becomes more frequent, indicating that
simply determining whether $\langle u_t, v_t\rangle < 0$ is insufficient to
characterize the effect of conflict.

Figure~\ref{fig:conflict_heatmap}(b) further shows the conflict strength
relative to the current generative velocity, measured as
$\|u_{t\parallel}\|_2 / \|v_t\|_2$.
ImageNet transformations exhibit strong early conflicts that rapidly decrease,
whereas CelebA-HQ \emph{Smile} and \emph{Age} edits start with weaker
conflicts, peak at intermediate stages, and then decline.
This reflects their different semantic changes: global ImageNet
transformations require stronger classifier correction early on, while
localized facial edits induce stronger conflict as the target attribute
emerges.
Despite these different dynamics, individual conflicts generally become
weaker toward the end of generation, as target semantics stabilize and less
classifier correction is required.

The different trends of conflict frequency and strength show that
directional conflict should not be treated as a binary event, but its
degree should also be considered.
As shown in Fig.~\ref{fig:conflict_heatmap}(c), Hard Projection and
Additive Guidance retain 0\% and 100\% of the conflicting component
respectively, whereas CLEAR adaptively retains 83.3--88.8\% across the
five tasks.

\begin{figure}[t]
    \centering
    \includegraphics[width=\columnwidth]{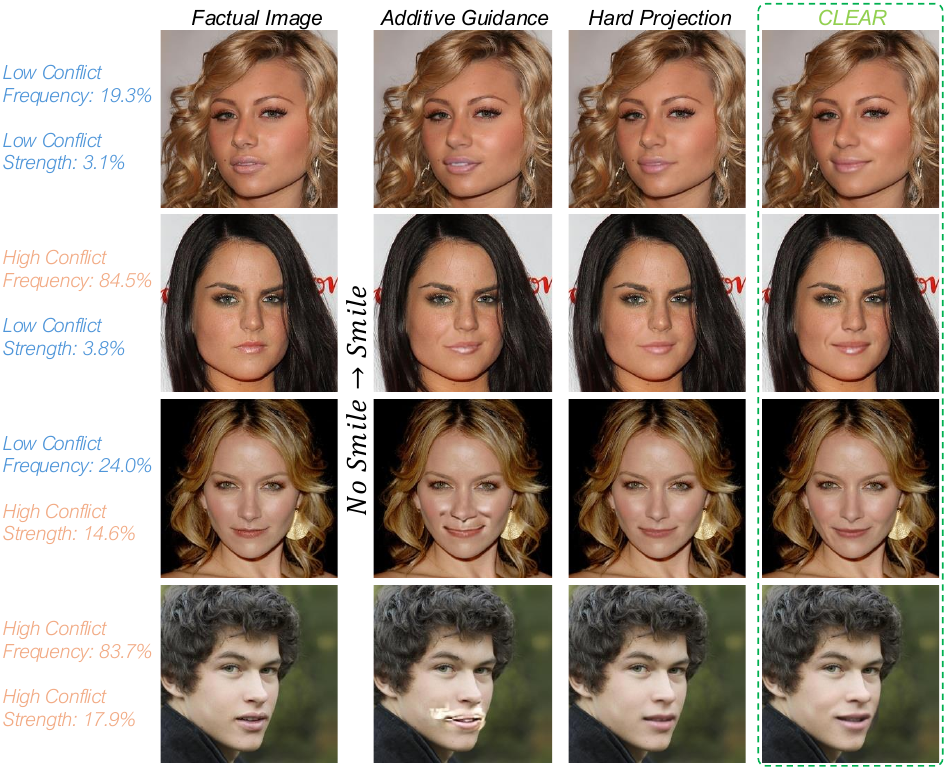}
\caption{
\textbf{Qualitative results under different conflict regimes.}
Rows show examples with different combinations of conflict frequency and strength. CLEAR consistently produces more natural and effective counterfactuals.
}
    \label{fig:conflict_vis}
\end{figure}

Figure~\ref{fig:conflict_vis} further illustrates the effect of different
conflict regimes on the final counterfactuals.
When conflicts are weak, even high conflict frequency leads to only limited
differences between guidance strategies.
In contrast, low-frequency but strong conflicts are already sufficient to
produce noticeable differences: Additive Guidance may cause excessive
modification, whereas Hard Projection may over-suppress useful classifier
guidance.
When both conflict frequency and strength are high, these differences become
most pronounced.
CLEAR instead adaptively attenuates the opposing component according to
conflict strength, achieving a better balance between target modification
and visual naturalness.
Additional qualitative results for the effects of conflict on  ImageNet are provided in the Appendix.

\subsection{Ablation Studies}
\label{sec:ablations}

We evaluate the main components of CLEAR on CelebA-HQ \emph{Smile}.
For Cooperative Flow Guidance, \emph{Additive Guidance} disables
conflict-aware attenuation by setting $\beta_t=0$, while
\emph{Hard Projection} completely removes the opposing component whenever
$\langle u_t,v_t\rangle<0$.
For Adaptive Support Preservation, \emph{Fixed Mask} keeps the initial
support unchanged, \emph{W/o Masked View} evaluates the classifier and
saliency directly on $x_t$, \emph{W/o Latent Mask} removes spatial gating
of the guided latent update, and \emph{W/o Pixel Reuse} omits the final
factual-pixel composition.

\textbf{Cooperative guidance.} Table~\ref{tab:ablation} confirms that both extremes are suboptimal:
fully retaining conflicting guidance degrades realism and non-target
preservation, while removing it entirely reduces counterfactual validity.
This supports CLEAR's central design of attenuating the conflicting
component according to its strength rather than treating conflict as a
binary event.

\textbf{Adaptive support preservation.}
Removing any of the three ASP components degrades the overall performance.
In particular, removing pixel reuse produces non-zero MSE outside the editable
support, revealing residual VAE reconstruction changes in the background.
CLEAR directly reuses factual pixels outside the final support, so zero
MSE in these regions is guaranteed by construction. This verifies the
composition operation, but does not independently validate support accuracy
or continuity at the composition boundary.
Further preservation visualizations and sensitivity analyses for
$\tau$, $t_s$, $N$, $s$, and $\rho$ are provided in the Appendix.

\begin{table}[t]
\centering
\small
\setlength{\tabcolsep}{2.0pt}
\renewcommand{\arraystretch}{1.05}
\caption{
\textbf{Main ablation on CelebA-HQ \emph{Smile}.}
Each variant changes only the specified component while keeping all
other settings fixed.
MSE is measured outside each variant's final editable support and is zero
by construction for variants using factual-pixel composition.
}
\label{tab:ablation}

\begin{tabularx}{\columnwidth}{
@{}
>{\raggedright\arraybackslash}p{0.30\columnwidth}
*{5}{>{\centering\arraybackslash}X}
@{}
}
\toprule
\textbf{Variant}
& FID$\downarrow$
& MNAC$\downarrow$
& COUT$\uparrow$
& FR(\%)$\uparrow$
& MSE$\downarrow$ \\
\midrule

\rowcolor{gray!10}
CLEAR
& 2.18
& 1.30
& 0.65
& 99.6
& 0.000 \\

\midrule
\multicolumn{6}{@{}l}{\textit{Cooperative guidance}} \\

Additive Guidance
& 4.47
& 2.16
& 0.71
& 99.7
& 0.000 \\

Hard Projection
& 3.66
& 1.63
& 0.64
& 95.6
& 0.000 \\

\midrule
\multicolumn{6}{@{}l}{\textit{Adaptive Support Preservation}} \\

Fixed Mask
& 3.40
& 1.57
& 0.61
& 93.7
& 0.000 \\

W/o Masked View
& 9.74
& 3.07
& 0.73
& 99.6
& 0.000 \\

W/o Latent Mask
& 6.42
& 2.48
& 0.58
& 92.8
& 0.000 \\

W/o Pixel Reuse
& 3.52
& 1.61
& 0.60
& 99.6
& 0.011 \\

\bottomrule
\end{tabularx}
\end{table}

\section{Conclusion and Limitations}
\label{sec:conclusion}

CLEAR improves latent visual counterfactual generation by addressing
directional conflicts between classifier guidance and the generative flow,
as well as reconstruction leakage outside the editable region.
Cooperative Flow Guidance uses the instantaneous Flow Matching direction
to preserve useful guidance while selectively attenuating components that
conflict with the generative process.
We further provide a novel way to systematically analyze directional conflicts
throughout counterfactual generation and clarify the interaction between
classifier guidance and the generative process.
Adaptive Support Preservation constrains guided updates in latent space and
directly preserves factual content outside the final editable region.
Experiments on CelebA-HQ and ImageNet show that CLEAR achieves high
counterfactual quality while retaining the efficiency of latent generation.

Our evaluation covers a limited range of counterfactual tasks, and the
quantitative ablation studies focus on CelebA-HQ Smile. Extending the
evaluation and ablations to more diverse tasks is a direction for future work.

\newpage
{
    \small
    \bibliographystyle{ieeenat_fullname}
    \bibliography{main}
}

\clearpage
\appendix
\renewcommand{\thefigure}{S\arabic{figure}}
\renewcommand{\thetable}{S\arabic{table}}
\renewcommand{\theequation}{S\arabic{equation}}
\setcounter{figure}{0}
\setcounter{table}{0}
\setcounter{equation}{0}

\section{Additional Analysis}

\subsection{Additional Qualitative Results under Different Conflict Regimes}

\begin{figure}[t]
    \centering
    \includegraphics[width=\columnwidth]{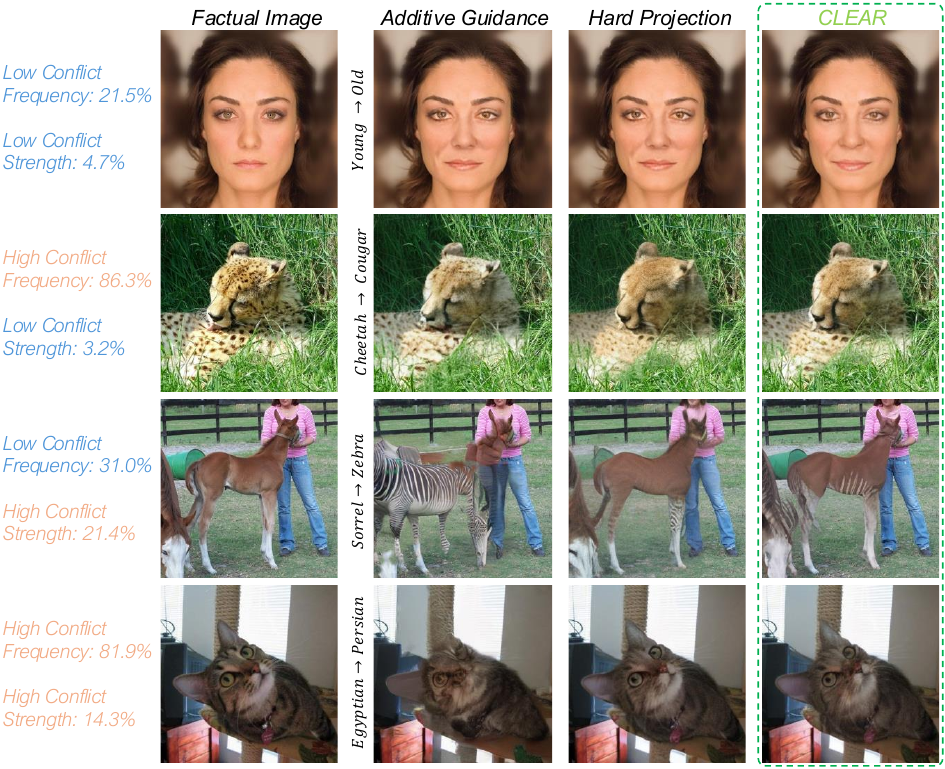}
    \caption{
    \textbf{Additional qualitative results under different conflict regimes.}
    Results on CelebA-HQ Age and ImageNet further demonstrate the consistent
    behavior of CLEAR across different conflict settings.
    }
    \label{fig:additional_conflict}
\end{figure}

Figure~\ref{fig:additional_conflict} extends the qualitative conflict analysis
in Fig.~8 of the main paper to CelebA-HQ \emph{Age} and ImageNet.
The examples cover different combinations of conflict frequency and strength
for both localized facial editing and global semantic transformations.
Consistent with the observations in the main paper, weak conflicts produce
relatively small differences between guidance strategies, whereas stronger
conflicts lead to more visible differences in the resulting counterfactuals.
Additive Guidance can introduce excessive modification, while Hard Projection
may suppress useful classifier guidance.
CLEAR instead adaptively attenuates the conflicting component according to
its strength, producing natural and effective counterfactuals across these
different settings.

\subsection{Adaptive Support Preservation}

\begin{figure}[t]
    \centering
    \includegraphics[width=\columnwidth]{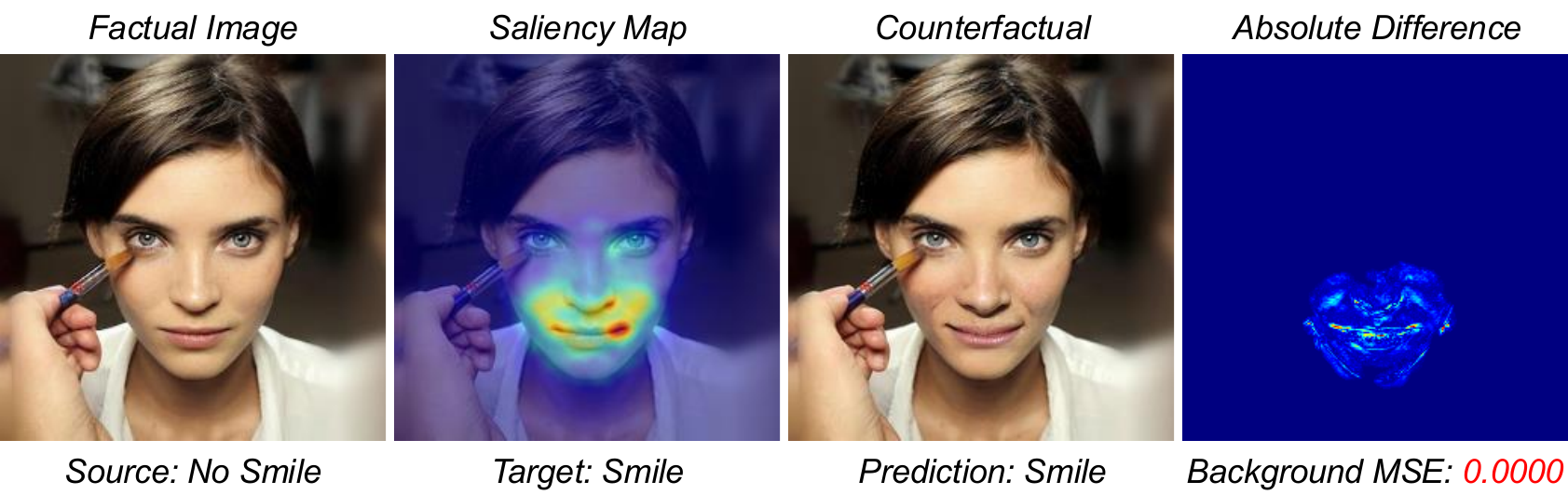}
    \caption{
    \textbf{Visualization of localized counterfactual editing.}
    The editable support concentrates changes in target-relevant regions while
    preserving the remaining factual content.
    }
    \label{fig:fig1}
\end{figure}

\begin{figure}[t]
    \centering
    \includegraphics[width=0.9\columnwidth]{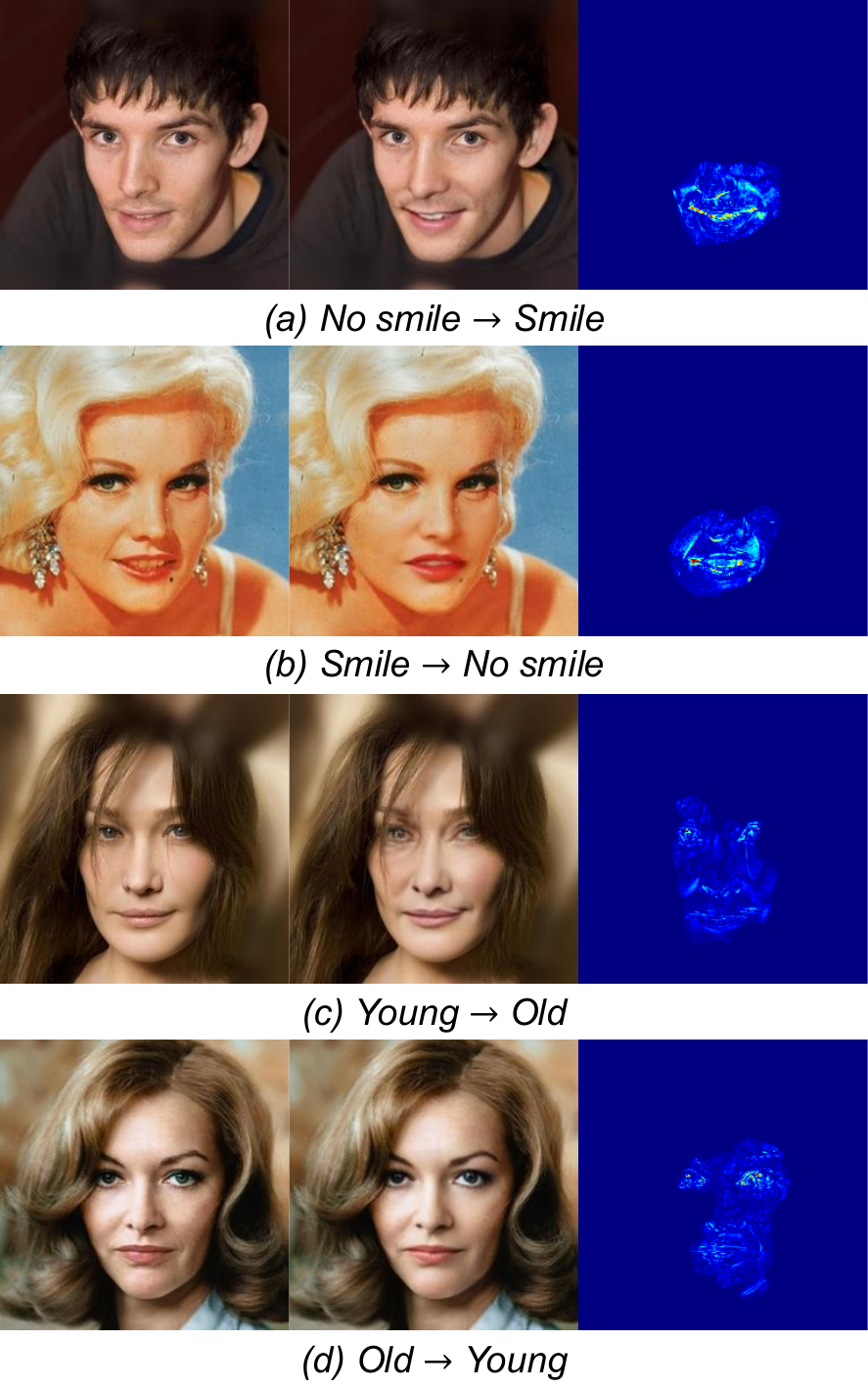}
    \caption{
    \textbf{Exact preservation with Adaptive Support Preservation.}
    From left to right: factual image, CLEAR counterfactual, and pixel-wise
    difference map. Pixels outside the editable support remain unchanged.
    }
    \label{fig:mse}
\end{figure}

Latent-space editing alone does not guarantee exact preservation of
decision-irrelevant content.
Even when classifier-driven updates are spatially restricted in latent space,
VAE decoding can still introduce reconstruction changes outside the intended
editable region.
Adaptive Support Preservation (ASP) addresses this issue at both the latent
and image levels.
During generation, it restricts classifier-driven updates to a dynamically
estimated editable support, while after generation it directly reuses factual
pixels outside the final support.

Figure~\ref{fig:fig1} visualizes the editable support together with the
resulting counterfactual changes.
The modifications closely follow the identified target-relevant regions,
showing that ASP localizes counterfactual editing rather than allowing
unnecessary changes across the image.
This is particularly important for latent VCEs, where restricting the latent
trajectory alone does not necessarily prevent reconstruction changes
introduced by the VAE decoder.

Figure~\ref{fig:mse} further illustrates the preservation behavior across
additional \emph{Smile} and \emph{Age} examples.
Although CLEAR performs counterfactual generation in latent space and decodes
the final latent through a VAE, pixels outside the final editable support are
taken directly from the factual image.
These regions are therefore exactly preserved and unaffected by VAE
reconstruction.
Compared with approximate preservation obtained only through latent
constraints, factual-pixel reuse provides exact pixel-level preservation outside the
estimated support. This guarantee does not establish that the support
correctly separates decision-relevant and decision-irrelevant content.

\subsection{Hyperparameter Sensitivity}

\begin{table}[t]
\centering
\small
\setlength{\tabcolsep}{2.0pt}
\renewcommand{\arraystretch}{1.05}
\caption{
\textbf{Hyperparameter sensitivity on CelebA-HQ \emph{Smile}.}
The default configuration uses $\tau=0.5$, $t_s=0.6$, $N=20$,
$s=8$, and $\rho=5\%$.
}
\label{tab:supp_sensitivity}

\begin{tabularx}{\columnwidth}{
@{}
>{\raggedright\arraybackslash}p{0.31\columnwidth}
*{4}{>{\centering\arraybackslash}X}
@{}
}
\toprule
\textbf{Setting}
& FID$\downarrow$
& MNAC$\downarrow$
& COUT$\uparrow$
& FR(\%)$\uparrow$ \\
\midrule

\rowcolor{gray!10}
Default
& 2.18
& 1.30
& 0.65
& 99.6 \\

\midrule
\multicolumn{5}{@{}l}{\textit{Conflict sensitivity}} \\

$\tau=0.25$
& 2.54
& 1.33
& 0.59
& 95.2 \\

$\tau=1.0$
& 4.28
& 1.71
& 0.68
& 99.7 \\

\midrule
\multicolumn{5}{@{}l}{\textit{Starting-time sensitivity}} \\

$t_s=0.4$
& 5.42
& 3.27
& 0.74
& 99.9 \\

$t_s=0.8$
& 3.40
& 1.42
& 0.65
& 86.4 \\

\midrule
\multicolumn{5}{@{}l}{\textit{Sampling-step sensitivity}} \\

$N=10$
& 2.98
& 1.35
& 0.57
& 90.1 \\

$N=40$
& 3.66
& 1.51
& 0.68
& 99.5 \\

\midrule
\multicolumn{5}{@{}l}{\textit{Guidance-scale sensitivity}} \\

$s=4$
& 2.21
& 1.30
& 0.44
& 93.5 \\

$s=12$
& 5.82
& 3.40
& 0.72
& 99.7 \\

\midrule
\multicolumn{5}{@{}l}{\textit{Support-ratio sensitivity}} \\

$\rho=3\%$
& 2.12
& 1.29
& 0.42
& 88.0 \\

$\rho=8\%$
& 2.63
& 1.48
& 0.67
& 99.6 \\

\bottomrule
\end{tabularx}
\end{table}

Table~\ref{tab:supp_sensitivity} evaluates the sensitivity of CLEAR to
$\tau$, $t_s$, $N$, $s$, and $\rho$ on CelebA-HQ \emph{Smile}, varying one
parameter at a time from the default configuration.
The results show substantial trade-offs between counterfactual validity,
realism, and preservation, indicating that parameter selection matters.
This single-task analysis does not establish robustness across datasets.

The conflict threshold $\tau$ controls how strongly the conflicting
classifier component is attenuated.
A smaller value ($\tau=0.25$) applies stronger attenuation and reduces the
flip rate from 99.6\% to 95.2\%.
Conversely, $\tau=1.0$ retains more classifier influence and reaches 99.7\%
flip rate, but increases FID and MNAC to 4.28 and 1.71, respectively.
This supports moderate conflict-dependent attenuation rather than either
aggressively suppressing or largely retaining conflicting guidance.

The starting time $t_s$ determines how much generative freedom is available
for counterfactual modification.
Starting earlier ($t_s=0.4$) produces strong validity
(FR $=99.9\%$, COUT $=0.74$), but substantially worsens FID and MNAC.
Starting later ($t_s=0.8$) keeps the trajectory closer to the factual image,
but restricts the target edit and reduces the flip rate to 86.4\%.
The default $t_s=0.6$ therefore provides a better balance between sufficient
editing freedom and factual preservation.

A similar trade-off appears for the number of sampling steps $N$.
Using only $N=10$ limits classifier-driven modification and reduces the
flip rate to 90.1\%, whereas increasing to $N=40$ restores high validity but
does not improve realism or preservation over the default $N=20$.
Additional sampling steps therefore provide limited benefit once sufficient
guidance has been applied.

The guidance scale $s$ controls the maximum strength of classifier guidance.
With $s=4$, weaker guidance preserves realism but produces insufficient
target modification, as reflected by COUT $=0.44$ and FR $=93.5\%$.
Increasing the scale to $s=12$ improves validity
(COUT $=0.72$, FR $=99.7\%$), but considerably degrades FID and MNAC.
This behavior is consistent with the motivation of Cooperative Flow Guidance:
classifier guidance should be strong enough to achieve the target change
without overwhelming the generative trajectory.

Finally, the support ratio $\rho$ controls the spatial extent of
counterfactual editing.
A small support ($\rho=3\%$) achieves low FID and MNAC but restricts the
editable region too strongly, reducing the flip rate to 88.0\%.
Increasing the support to $\rho=8\%$ restores high validity, but slightly
increases FID and MNAC as modifications are allowed over a larger region.
The default $\rho=5\%$ provides a balanced compromise between localized
editing and successful counterfactual generation.

Taken together, these results show that the default configuration provides
a consistent balance between validity, realism, and preservation.
The observed sensitivity trends also agree with the intended roles of the two
main components of CLEAR: Cooperative Flow Guidance controls \emph{how
strongly} classifier information modifies the trajectory, while Adaptive
Support Preservation controls \emph{where} those modifications are allowed.

\end{document}